\documentclass[conference]{IEEEtran}
\IEEEoverridecommandlockouts
\usepackage{cite}
\usepackage{amsmath,amssymb,amsfonts}
\usepackage{algorithmic}
\usepackage{graphicx}
\usepackage{textcomp}
\usepackage{xcolor}
\usepackage{booktabs}
\usepackage{tabularx}
\usepackage{array}
\usepackage{mathrsfs}
\def\BibTeX{{\rm B\kern-.05em{\sc i\kern-.025em b}\kern-.08em
    T\kern-.1667em\lower.7ex\hbox{E}\kern-.125emX}}
\begin{document}

\title{Learning Thermal-Aware Locomotion Policies for an Electrically-Actuated Quadruped Robot}






\author{%
Letian Qian, Yuhang Wan, Shuhan Wang, Xin Luo\textsuperscript{*} \\
\textit{State Key Laboratory of Intelligent Manufacturing Equipment and Technology} \\
\textit{Huazhong University of Science and Technology} \\
Wuhan, China \\
\{qianletian93, m202470537\}@hust.edu.cn, wangshuhan1995@gmail.com, mexinluo@hust.edu.cn%
\thanks{\textsuperscript{*}Corresponding author.}%
}

\maketitle

\begin{abstract}
 Electrically-actuated quadrupedal robots possess high mobility on complex terrains, but their motors tend to accumulate heat under high-torque cyclic loads, potentially triggering overheat protection and limiting long-duration tasks. This work proposes a thermal-aware control method that incorporates motor temperatures into reinforcement learning locomotion policies and introduces thermal-constraint rewards to prevent temperature exceedance. Real-world experiments on the Unitree A1 demonstrate that, under a fixed 3 kg payload, the baseline policy triggers overheat protection and stops within approximately 7 minutes, whereas the proposed method can operate continuously for over 27 minutes without thermal interruptions while maintaining comparable command-tracking performance, thereby enhancing sustainable operational capability.
\end{abstract}


\section{Introduction}

Quadrupedal robots exhibit strong terrain adaptability in scenarios such as inspection, transport, and disaster response, making them ideal candidates for continuous-operation tasks \cite{b20, b21}. However, their motors typically operate under cyclic high-torque loads, and factors such as load variations, elevated ambient temperatures, and limited cooling can cause motor heating to deviate significantly from design assumptions. Although nominal payloads and endurance specifications are usually provided during the robot design phase, these metrics are often based on ideal cooling conditions and moderate levels of locomotion. In real-world tasks, motor temperatures can quickly accumulate and trigger overheat protection, leading to performance degradation or task interruption. This issue becomes particularly pronounced under long-duration tasks, high payloads, or harsh outdoor conditions, posing a key limitation to the sustainable locomotion of quadrupedal robots.

Existing approaches for addressing motor thermal limitations can be broadly divided into hardware-level thermal capability enhancement and control-layer thermal constraint handling. The former aims to increase a motor’s allowable thermal envelope—e.g., by improving heat dissipation or increasing thermal mass/capacity—to delay the onset of overheating and sustain higher torque output \cite{b1, b2, b3}. However, such approaches do not exploit the intrinsic coupling between thermal dynamics and power delivery, and thus cannot proactively determine thermal violations under specific task loads or environmental conditions. As a result, protective control strategies remain necessary to guarantee continuous operation. In contrast, control-layer approaches incorporate thermal models or temperature feedback into cost functions or safety constraints, enabling policies or controllers to actively satisfy thermal safety requirements during execution \cite{b4}. These studies mainly target fixed-base systems, such as robotic arms, where thermal constraints typically apply to static or slowly varying trajectories and can be enforced simply by limiting torque or power. In quadrupedal robots, however, torque and power output not only influence motor temperature but are also critical for maintaining dynamic balance, foot-ground contact, and task execution. Consequently, thermal management and stability control are tightly coupled, and there is currently a lack of effective methods for proactively preventing motor overheating in legged robots.

To the end, this work proposes a thermal-aware locomotion control method for quadrupedal robots. The method incorporates motor temperatures and their dynamic evolution into the state space of a reinforcement learning policy and designs thermal-constraint rewards to encourage the policy to adapt its behavior when approaching thermal limits, thereby delaying the activation of overheat protection. The approach does not rely on additional hardware cooling and can accommodate sustainable operation under varying task and thermal conditions. We validate the proposed method in both simulation and on the real Unitree A1 platform. Results show that the method significantly reduces motor overheating risk while maintaining comparable command-tracking performance and extends continuous operation under fixed-load tasks from approximately 7 minutes to over 27 minutes without thermal interruptions, demonstrating the potential of proactive thermal management to enhance the endurance of quadrupedal robots.

\section{Motor Thermal Model for Quadruped Robots}

\subsection{Single-Motor Thermal Model}

The temperature of a single motor can be effectively predicated by a first-order thermal model \cite{b5}. The temperature evolution of the motor is mainly attributed to the Joule heating induced by the winding current and the thermal interaction with the ambient environment. According to \cite{b6}, the thermal behavior of the motor can be described by a single time constant thermal equation, representing the thermodynamic behavior of a homogeneous body at rest heated by electric current:

\begin{equation}
C_{\mathrm{M} \sim\mathrm{E}} \dot{T}=-\frac{T-T_\mathrm{E}}{R_{\mathrm{M}\sim\mathrm{E}}}+I^{2} R_{d}\label{eq：equaton1}
\end{equation}

\noindent{where} $T$ is the motor temperature, $T_E$ is the ambient temperature, 
$C_{M\sim E}$ and $R_{M\sim E}$ are the equivalent thermal capacitance and thermal resistance between the motor and the environment, 
and $R_d$ is the equivalent electrical resistance of the motor windings.

\begin{figure}[htbp]       
  \centering                
  \includegraphics[width=0.8\linewidth]{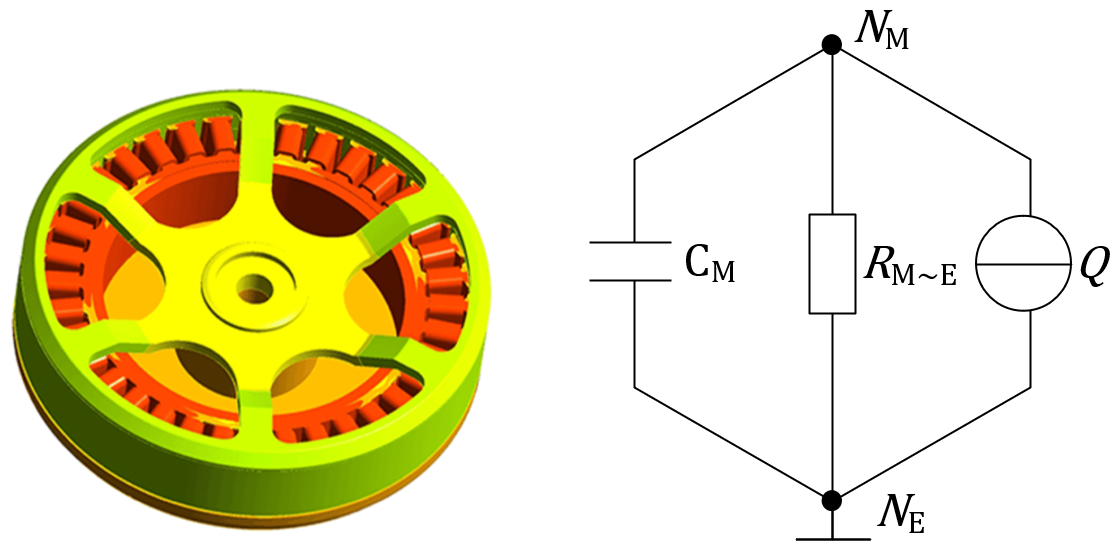}  
  \caption{Single-Motor Thermal Model.}          
  \label{fig:figure1}       
\end{figure}

\subsection{Whole-body Thermal Model of the Quadruped Robot}

\begin{figure}[htbp]       
  \centering                
  \includegraphics[width=0.95\linewidth]{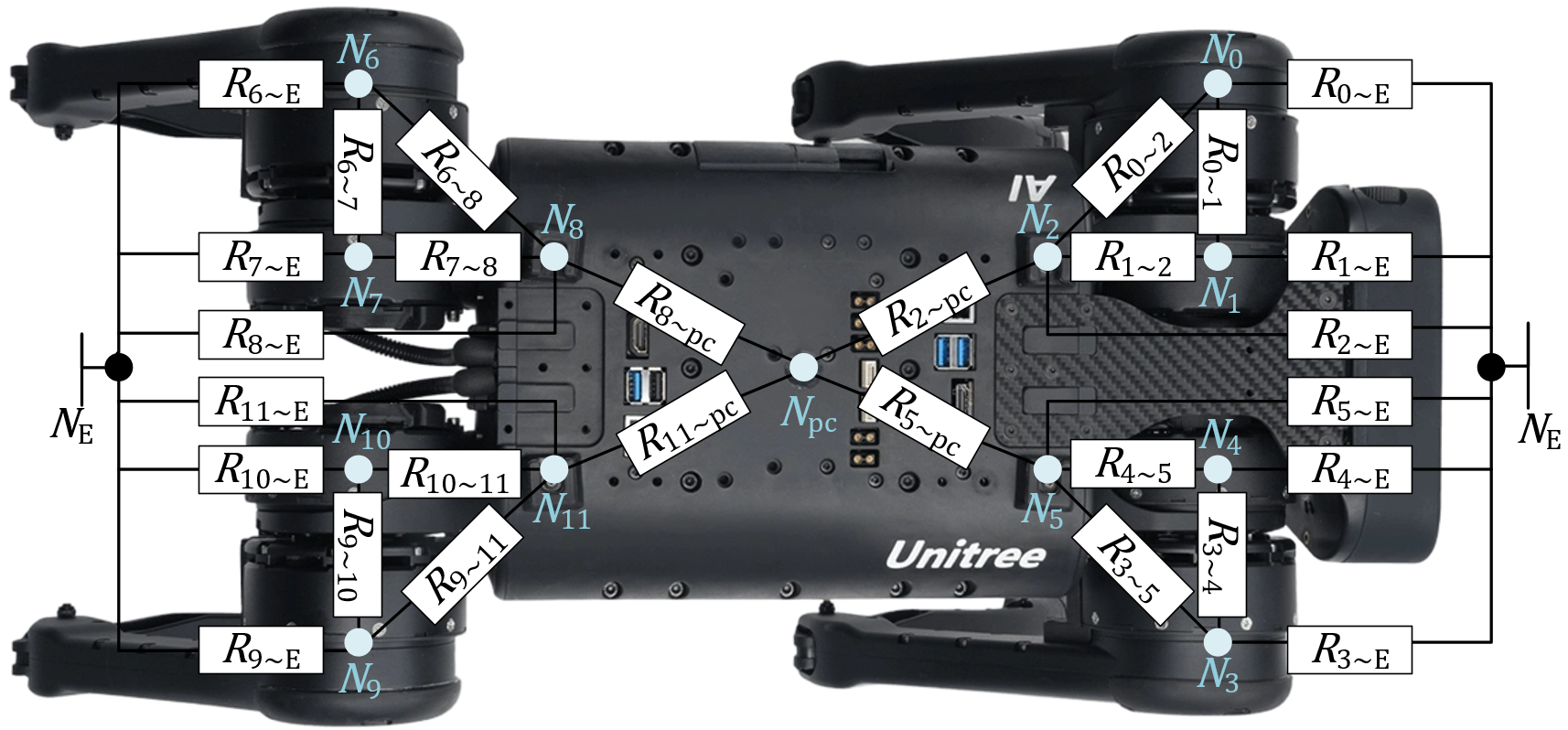}  
  \caption{Whole-body thermal model of the quadruped robot.}          
  \label{fig:figure2}       
\end{figure}

Due to the compact structure of quadruped robots, joint motors are densely distributed in space, resulting in significant thermal coupling effects between motors. Therefore, modeling each motor independently as a first-order thermal system is difficult to accurately predicate the temperature of all motors in the quadruped robot during locomotion. According to \cite{b7}, we construct a whole-body thermal model of the quadruped robot, as shown in Fig.~\ref{fig:figure2}.

$N_i$ represents the $i$-th motor corresponding to the single-motor thermal model shown in Fig.~\ref{fig:figure1}, serving as a thermal node. $N_{pc}$ represents the thermal node associated with the onboard computer. $R_{i\sim j}$ represents the equivalent thermal resistance between node $i$ and node $j$. The heat transferred between neighboring nodes can be written as:

\begin{equation}
Q_{i \sim j}=-\frac{T_{i}-T_{j}}{R_{i \sim j}}\label{eq：equaton2}
\end{equation}

For any thermal node $N_{i}$, considering its thermal interactions with neighboring nodes, its temperature can be modeled as:

\begin{equation}
C_{i\sim\mathrm{E}}\dot{T}_{i} = -\sum_{j\in\mathcal{M}(i)}\frac{T_{i}-T_{j}}{R_{i\sim j}} + I^2 R_d + Q_{i,\mathrm{other}}\label{eq：equaton3}
\end{equation}

\noindent{where} $\mathcal{M}(i)$ is the set of all neighboring nodes thermally connected to node $N_i$, and $Q_{i,\mathrm{other}}$ is additional heat generated by joint friction and the constant heat from the motor driver. Detailed modeling procedures can be found in \cite{b7}.

The whole-body thermal model of the quadruped robot can be formulated in state-space form, and the continuous-time system is further discretized as follows:

\begin{equation}
\boldsymbol{x}(k+1) = \boldsymbol{A}(h) \boldsymbol{x}(k+1) + \boldsymbol{B}(h) \boldsymbol{u}(k+1)\label{eq：equaton4}
\end{equation}

\noindent{where} $\boldsymbol{x} \in \mathbb{R}^{14 \times 1}$ is the temperatures of all nodes including motors, computer, and environment, and $\boldsymbol{u} \in \mathbb{R}^{14 \times 1}$ is the generated heat at each node, defined as $\boldsymbol{Q}_{\text{in}} = \boldsymbol{I}^2 \boldsymbol{R}_d + \boldsymbol{Q}_{i,\text{other}}$. $\boldsymbol{A}(h) \in \mathbb{R}^{14 \times 14}$ is the system matrix describing the thermal coupling between nodes, and $\boldsymbol{B}(h) \in \mathbb{R}^{14 \times 14}$ is the input matrix, $h$ is the sampling interval, and $k$ is the sampling index.

\section{Thermal-Aware Locomotion Policy Design}

\subsection{Training Framework}

\begin{figure}[htbp]       
  \centering                
  \includegraphics[width=1.0\linewidth]{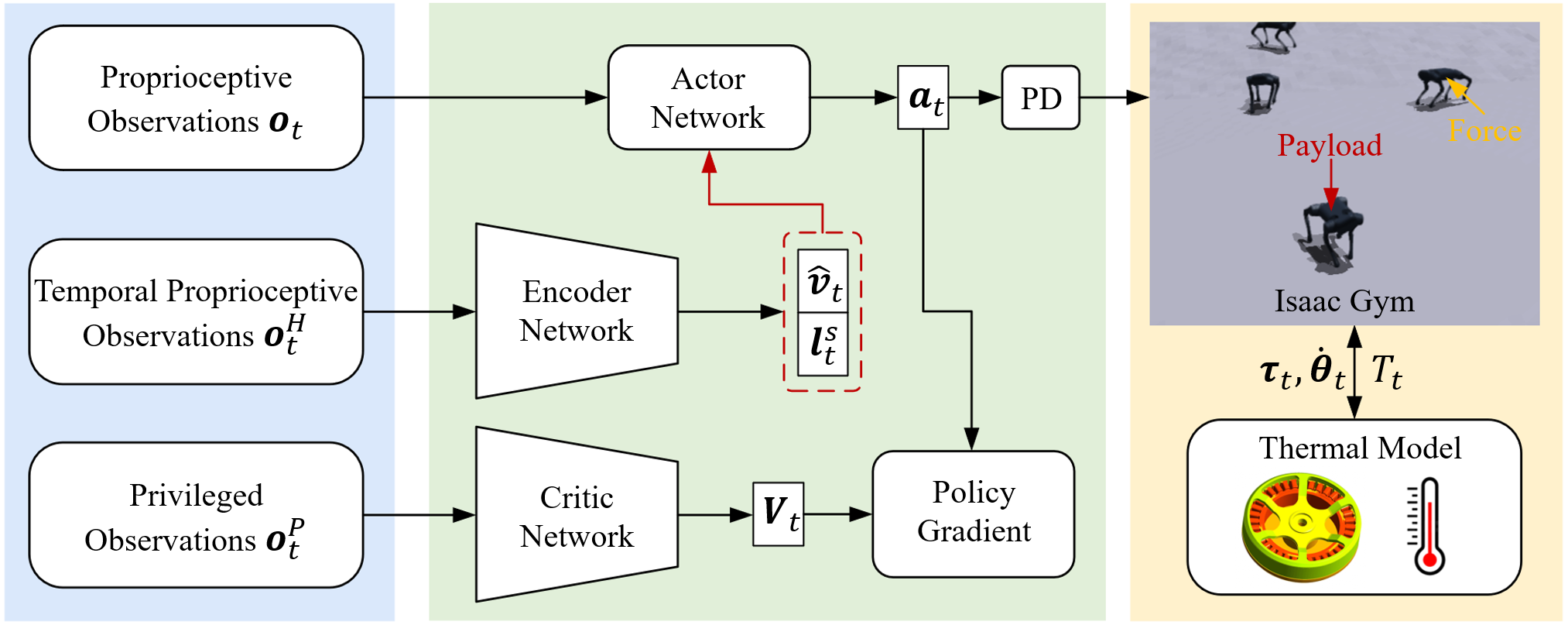}  
  \caption{Overview of the proposed training framework. During training, all observations are obtained from the simulator. Among them, the output frequency of all networks and the thermal model are both 50 Hz, while the PD controller outputs torques to the actuators at 200 Hz within the simulation.}          
  \label{fig:figure3}       
\end{figure}

According to prior reinforcement learning studies on quadrupedal locomotion control \cite{b8, b9}, we establish a policy training framework for motor heat management, as illustrated in Fig.~\ref{fig:figure3}. The Actor outputs an action $\boldsymbol{a}_t \in \mathbb{R}^{12}$ based on the robot’s state and its task which is added to the nominal joint angles $\boldsymbol{\theta}_0 \in \mathbb{R}^{12}$ to determine the target joint positions. Then target positions are fed into joint PD controllers to generate either the joint torques in the simulation or the motor currents for the real robot, thereby achieving quadrupedal locomotion. During training, the optimization process consists of two stages: Hybrid Internal Optimization (HIO), which updates the encoder network, and Proximal Policy Optimization (PPO) \cite{b10}, which optimizes the Actor and Critic network.


In real-world applications, it is difficult to access certain state information including linear velocity, and measurements obtained from the IMU and motor encoders are corrupted by noise. Therefore, quadrupedal locomotion is naturally formulated as a Partially Observable Markov Decision Process (POMDP). To improve value estimation during training and ensure policy convergence, we adopt an asymmetric Actor–Critic framework. In the simulation training phase, the Critic receives additional privileged observations $\boldsymbol{o}_t^p$ including linear velocity $\boldsymbol{v} \in \mathbb{R}^{3}$ and external force $\boldsymbol{F} \in \mathbb{R}^{3}$, in addition to the proprioceptive observation $\boldsymbol{o}_t$. The specific expression is as follows:

\begin{equation}
\begin{aligned}
\boldsymbol{o}_t &= \big( \boldsymbol{v}_t^{\mathrm{cmd}}, \boldsymbol{\omega}_t, \boldsymbol{g}_t, \boldsymbol{\theta}_t, \boldsymbol{\dot{\theta}}_t, \boldsymbol{T}_t, \boldsymbol{a}_{t-1} \big) \\
\boldsymbol{o}_t^p &= \big( \boldsymbol{v}_t, \boldsymbol{F}_t \big)
\end{aligned}
\label{eq:equation5}
\end{equation}

\noindent{where} $\boldsymbol{v}_t^{\mathrm{cmd}} \in \mathbb{R}^{3}$ denotes the commanded velocity in the robot’s base frame. $\boldsymbol{\omega}_t \in \mathbb{R}^{3}$ and $\boldsymbol{g}_t \in \mathbb{R}^{3}$ denote the angular velocity and gravity vector in the robot’s base frame. $\boldsymbol{\theta}_t \in \mathbb{R}^{12}$ , $\boldsymbol{\dot{\theta}}_t \in \mathbb{R}^{12}$ and $\boldsymbol{T}_t \in \mathbb{R}^{12}$ denote the motor positions, velocities and temperatures. $\boldsymbol{a}_{t-1} \in \mathbb{R}^{12}$ denotes the previous action.

In addition to the proprioceptive observation $\boldsymbol{o}_t$, the observation for the Actor also includes the estimated velocity $\hat{\boldsymbol{v}}_t$ and the latent feature vector $\boldsymbol{l}_t^{s} \in \mathbb{R}^{16}$ with the encoder using the previous six frames of proprioceptive observations to generate them. Detailed implementation can be found in \cite{b9}.

\subsection{Temperature Randomization}

We incorporate the whole-body thermal model of the quadruped robot into the simulation to update motor temperatures of all agents in real time. Since torque $\tau$ is approximately linearly proportional to current $I$, we use the motor torque $\tau$ as a substitute for the motor current $I$. Meanwhile, the torque varies at a much higher frequency than the update rate of the thermal model, we therefore use the root-mean-square (RMS) of the torque derived from Joule heating theory as the effective thermal input to ensure modeling accuracy:

\begin{equation}
\bar{\tau}_t = \sqrt{\frac{1}{n} \sum_{m=0}^{n} \tau_{m,\mathrm{sample}}^{2}}\label{eq：equaton6}
\end{equation}

\noindent{where} $\tau_{m,\mathrm{sample}}$ denotes the $m$-th torque sample within the thermal model update interval. The thermal model parameters for the Unitree A1 are taken from \cite{b7}, and the thermal model is updated at the same frequency as the Actor outputs.

To reduce the sim-to-real gap, we apply domain randomization during training using the parameters listed in Table.~\ref{tab:table1} to initialize the robot states. Taking the limited episode length of each agent into consideration, we initialize motor temperature within the range $\left[ T_{\max} - 35,\; T_{\max} + 10 \right]$, ensuring a greater number of samples near $T_{\max}$ at the beginning to improve the efficiency of policy learning. Furthermore, under conditions without payload or persistent external forces, motor temperatures rarely reach the overheat threshold. To construct operating conditions with realistic motor overheat risk, we additionally introduce domain randomization of eccentric payloads and persistent external forces.


\renewcommand{\arraystretch}{1.5}
\begin{table}[htbp]
\centering
\caption{Domain Randomizations and Respective Range}
\label{tab:table1}
\begin{tabular}{ccc}
\toprule
\textbf{Parameters} & \textbf{Range [Min, Max]} \\
\midrule
Payload Mass & $[0, 4]$ \\

CoM Displacement & 
$[-0.1, 0.1] \times [-0.1, 0.1] \times [-0.1, 0.1]$ \\

External Force & 
$[-30, 30] \times [-30, 30] \times [-30, 30]$ \\

Ground Friction & $[0.2, 1.25]$ \\

Initial Joint Positions & 
$[0.5, 1.5] \times \text{nominal value}$ \\

System Delay & $[0, 3\Delta t]$ \\

Motor Strength & 
$[0.8, 1.2] \times \text{motor torque}$ \\

Initial Motor Temperature & 
$[T_{\max}-25,\, T_{\max}+10]$ \\

Environment Temperature & $[0, 35]$\\
\bottomrule
\end{tabular}
\end{table}

\subsection{Reward Function}

To guide the Actor in learning a locomotion policy with motor heat management, we design a reward function related to motor temperature that encourages the robot to reduce heat generation of overheated motors. Due to the strong inertia of the whole-body thermal model, motor temperatures cannot decrease significantly within an episode. Following \cite{b11}, we convert the inequality constraint requiring the motor temperature to remain below $T_{\max}$ into a Control Barrier Function (CBF) \cite{b12}, which improves the responsiveness of the reward to actions outputted by the Actor:

\begin{equation}
- \dot{T}_t + \gamma_T \left( T_{\max} - T_t \right) \ge 0, \quad \gamma_T > 0\label{eq：equaton7}
\end{equation}

The CBF condition effectively imposes a constraint on the motor temperature derivative: when the temperature approaches or exceeds the threshold, the system is required to satisfy 
$\dot{T} \le 0$ to prevent further heating. However, within an episode, 
$\gamma_T \left( T_{\max} - T_t \right)$ typically exhibits limited variation, meaning that the reward can still be heavily influenced by the initial temperature. 
To mitigate this initialization bias and make the reward more focused on the motors whose temperature is near the threshold, we introduce a clipped temperature $\boldsymbol{T}_t^{\mathrm{clip}} = \operatorname{clip}\big(\boldsymbol{T}_t,\, {T}_{\min}^{\mathrm{clip}},\, {T}_{\max}^{\mathrm{clip}}\big)$ and compute penalties only based on violations of the CBF constraints for each motor. All rewards and their weights are summarized in Table~\ref{tab:table2}.

\renewcommand{\arraystretch}{1.5}
\begin{table}[htbp]
\centering
\caption{Rewards for Thermal-Aware Locomotion Policy}
\label{tab:table2}
\small 
\begin{tabularx}{\columnwidth}{
  >{\centering\arraybackslash}m{0.28\columnwidth}
  >{\centering\arraybackslash}m{0.5\columnwidth}
  >{\centering\arraybackslash}m{0.12\columnwidth}
}
\toprule
\textbf{Reward Item} & \textbf{Equation} & \textbf{Weight} \\
\midrule

Linear velocity tracking &
$\exp\!\Big(-\frac{\|\boldsymbol{v}_{xy}^{\mathrm{cmd}} - \boldsymbol{v}_{xy}\|_2^2}{\sigma}\Big)$ & 1.0 \\

Angular velocity tracking &
$\exp\!\Big(-\frac{(\omega_{\mathrm{yaw}}^{\mathrm{cmd}} - \omega_{\mathrm{yaw}})^2}{\sigma}\Big)$ & 0.8 \\

Linear velocity (z) & $v_z^2$ & -2.0 \\

Angular velocity (roll-pitch) & $\boldsymbol{\omega_{xy}}^2$ & -0.05 \\

Orientation & $\|\mathbf{g}\|_2^2$ & -0.2 \\

Joint accelerations & $\|\ddot{\boldsymbol{\theta}}\|_2^2$ & -2.5e-7 \\

Termination & / & -200 \\

Body height & $(h^{\mathrm{target}} - h)^2$ & -1.0 \\

Foot clearance &
$\sum_{i=0}^{3} ( p_z^{\mathrm{target}} - p_z^i )^2 \cdot v_{xy}^i$ & -0.01 \\

Action rate &
$\|\boldsymbol{a}_t - \boldsymbol{a}_{t-1}\|_2^2$ & -0.01 \\

Smoothness &
$\|\boldsymbol{a}_t - 2\boldsymbol{a}_{t-1} + \boldsymbol{a}_{t-2}\|_2^2$ & -0.01 \\

Motor temperature &
$- \|\min( -\dot{\boldsymbol{T}}_t + \gamma_T (\boldsymbol{T}_{\max} - \boldsymbol{T}_t^{\mathrm{clip}}), 0 )\|_1$ & 2.0 \\

\bottomrule
\end{tabularx}
\end{table}

To ensure that the CBF constraint remains satisfied even when a high-temperature motor receives zero input, we determine the weight coefficient $\gamma_T$ based on \eqref{eq：equaton4}:

\begin{equation}
\frac{[\boldsymbol{A}(h)- \boldsymbol{I}]\, \boldsymbol{T}_{\max}^{\text{clip}}}{h}+\gamma_T\bigl(\boldsymbol{T}_{\max}-\boldsymbol{T}_{\max}^{\text{clip}}\bigr)\ge0\label{eq：equaton8}
\end{equation}

\noindent{where} $\boldsymbol{T}_{\max}^{\text{clip}}$ denotes a temperature vector in which all nodes except the environmental node are set to ${T}_{\max}^{\mathrm{clip}}$. All parameters of rewards are reported in Table ~\ref{tab:table3}.




\begin{table}[t]
\centering
\caption{Reward Parameters}
\label{tab:table3}

\setlength{\tabcolsep}{4pt}
\renewcommand{\arraystretch}{1.2}

\begin{tabular}{p{0.22\columnwidth} c | p{0.22\columnwidth} c}
\hline
\multicolumn{2}{c|}{\textbf{Motor Temperature Reward Function}} 
& \multicolumn{2}{c}{\textbf{Other Reward Function}} \\
\hline
$T_{\max}$ & $60^\circ\mathrm{C}$ 
& $\sigma$ & 0.25 \\

$[T_{\min}^{\mathrm{clip}},\, T_{\max}^{\mathrm{clip}}]$ 
& $[55^\circ\mathrm{C},\, 65^\circ\mathrm{C}]$ 
& $h^{\mathrm{target}}$ & 0.3\,m \\

$\gamma_T$ & 0.35 
& $p_z^{\mathrm{target}}$ & -0.2\,m \\
\hline
\end{tabular}
\end{table}

\section{EXPERIMENTS AND RESULTS}
\subsection{Experimental Setup}
We used the Isaac Gym simulator to build the simulation environment for training, and deployed  8192 robot agents in parallel to concurrently learn for 6000 epochs with the episode length of each agent set to 100 time steps(20 s). Given our focus on the management of motor temperature during quadrupedal locomotion, we trained the policies to walk omnidirectionally on flat ground. To enhance policy stability and robustness, We applied uniform random noise ranging from -2 cm to +2 cm to the terrain, which included light slopes up to a 2\% incline \cite{b13}. For sim-to-sim validation, we used Gazebo as the simulation environment, which allows direct and real-time access to accurate measurements such as velocity, GRFs(ground reaction forces), and base height for detailed experimental analysis. In real-world experiments, the trained encoder network shown in Fig.~\ref{fig:figure3} was used to estimate the base velocity, and temperatures feedback from motors were directly fed into the policy as inputs.

\subsection{Real-world Locomotion Results}
\begin{figure}[htbp]       
  \centering                
  \includegraphics[width=0.95\linewidth]{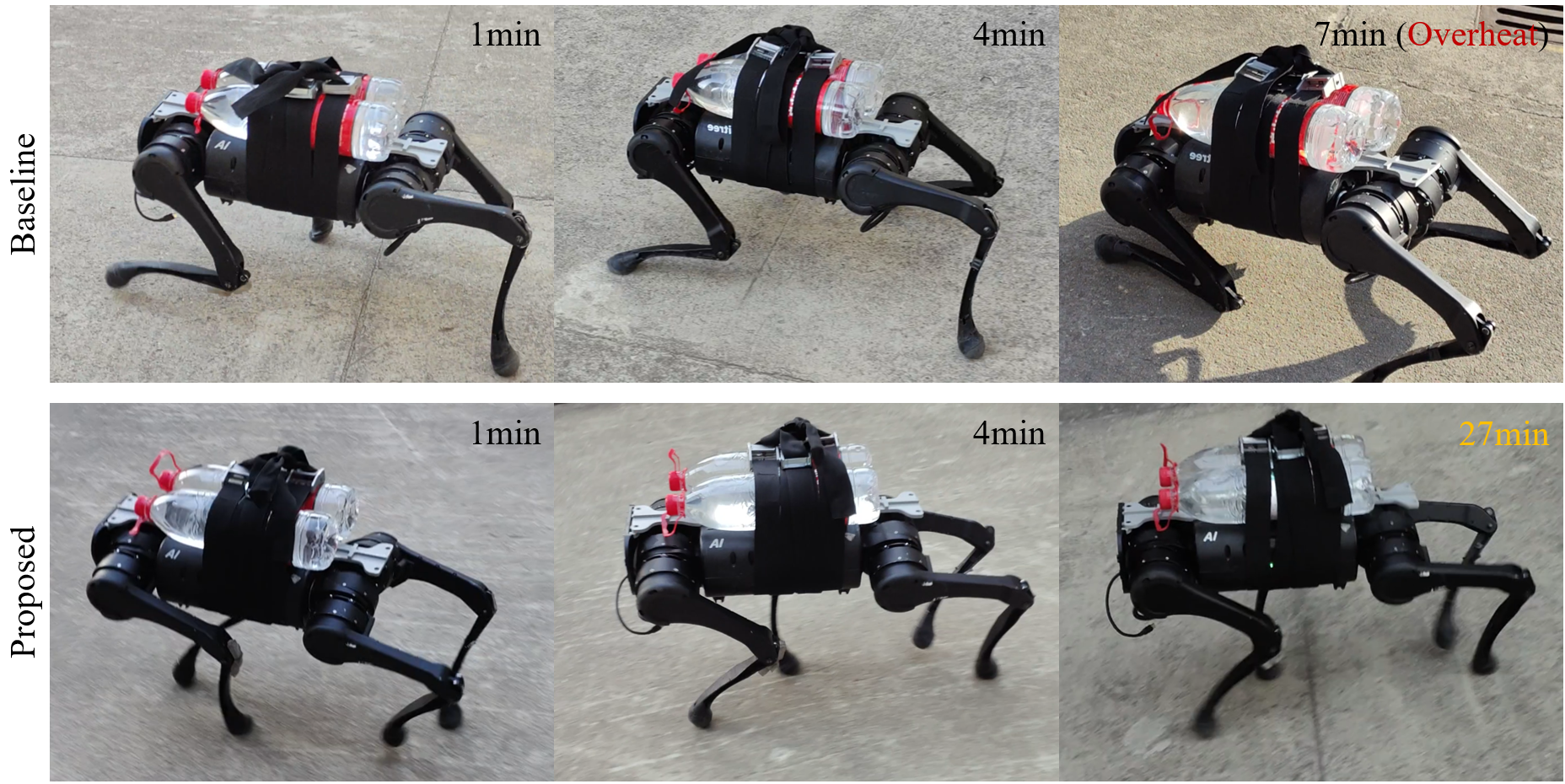}  
  \caption{Comparison between baseline and proposed policy for continuous locomotion.}          
  \label{fig:figure4}       
\end{figure}

\begin{figure}[htbp]       
  \centering                
  \includegraphics[width=0.95\linewidth]{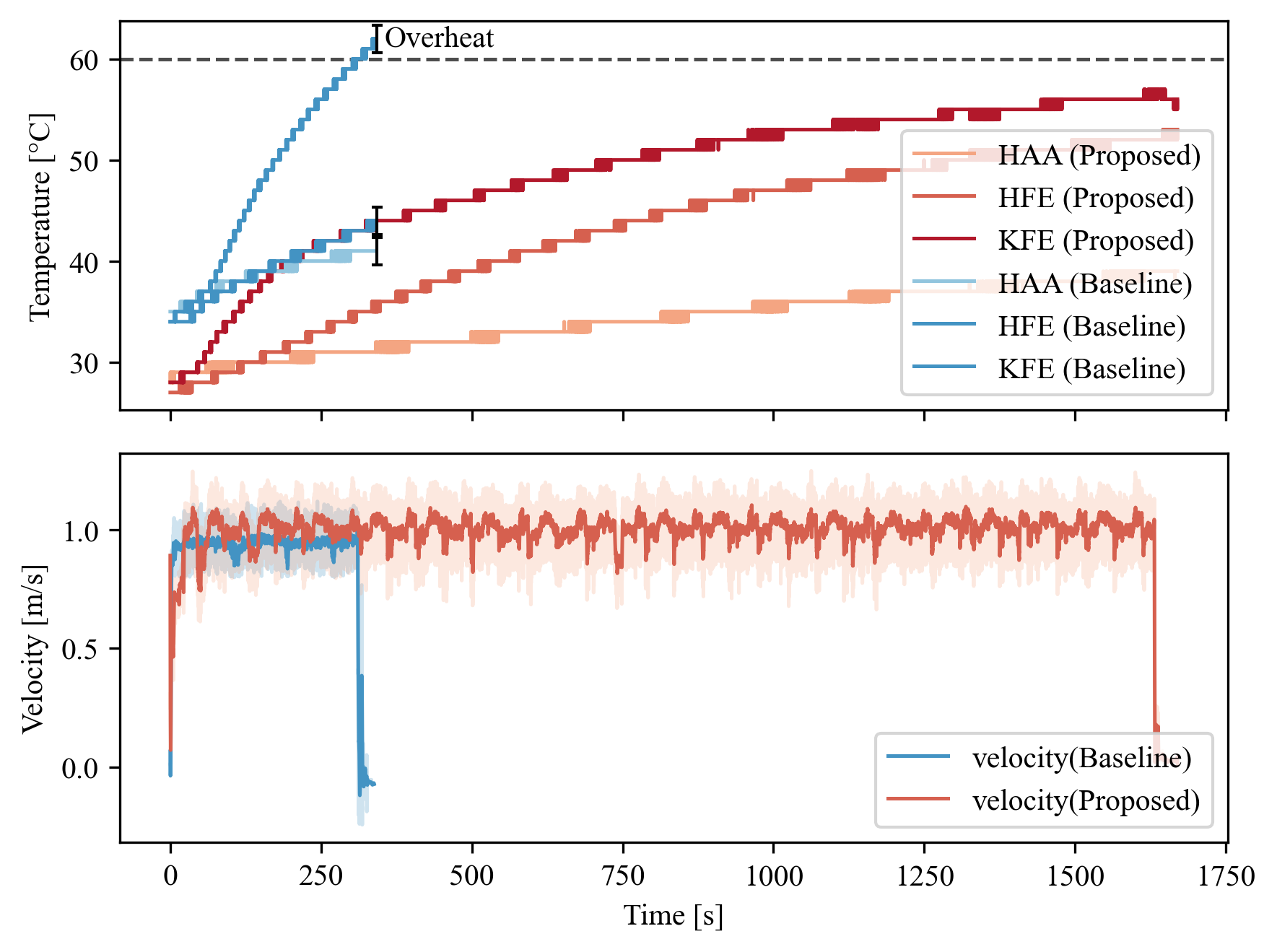}  
  \caption{Real-world experimental results. In the upper figure, the dashed line indicates the motor temperature threshold. Due to outdoor factors, it is not possible to ensure the same initial temperature in these experiments. The lower figure shows the robot’s base velocity under joystick control, which was estimated by the encoder network illustrated in Fig.~\ref{fig:figure3}.} 
  \label{fig:figure5}       
\end{figure}

To validate the effectiveness of our policy, we used a policy trained without the motor temperature reward as the baseline, and compared the motor temperature evolution under different strategies. In real-world experiments, a 3 kg external payload was mounted on the robot’s back to increase the risk of motor overheating. The robot was commanded to move forward at 1 m/s and walk continuously on flat terrain, as shown in Fig.~\ref{fig:figure4}. The curves in Fig.~\ref{fig:figure5} show that the baseline failed to continue walking in only 7 minutes due to overheating of the front-left knee motor. In contrast, controlled by the proposed policy, the robot walked stably under the same conditions until the battery was depleted. During more than 26 minutes of continuous locomotion, the temperatures of all motors remained below the threshold temperature ${T}_{\max}$. 

\begin{figure*}[htbp]       
  \centering                
  \includegraphics[width=0.8\linewidth]{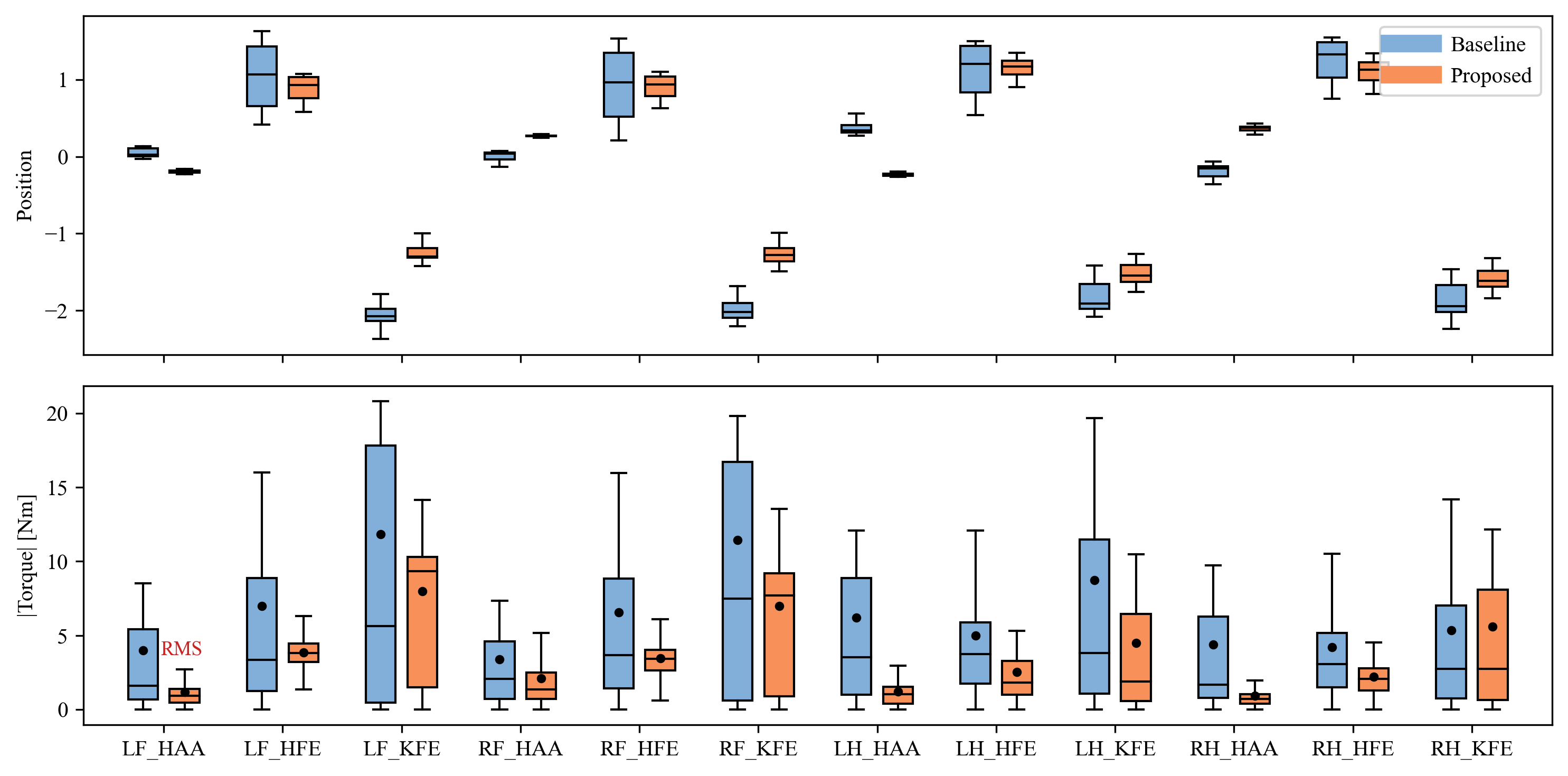}  
  \caption{Comparison of joint positions and torques of the robot during locomotion under the baseline and proposed policy. Torques are shown in absolute values, with black circles indicating the root-mean-square (RMS) values.} 
  \label{fig:figure6}       
\end{figure*}

\subsection{Analysis of Thermal-Aware Locomotion Performance}

The hardware experiments demonstrate that the proposed thermal-aware policy substantially mitigated motor temperature rise under load while preserving tracking performance. This section examines the underlying mechanism enabling this behavior.

Mechanism analysis requires access to joint torques, three-dimensional ground reaction forces, joint-to-foot Jacobians, and body posture. Since accurate 3-D force measurements are difficult to obtain on hardware, the analysis was conducted in simulation. To accentuate the differences between controllers, a more challenging condition was considered by applying an additional 4 kg lateral load to the upper-left region of the torso while commanding forward locomotion at $v_x^\ast = 1.0\,\mathrm{m/s}$. This configuration increased the coupling between posture regulation and load distribution and facilitates observation of the resulting compensatory behaviors.

Fig.\ref{fig:figure6} reports joint-space statistics. Under the proposed controller, joint angle excursions were reduced, and both peak and RMS joint torques were substantially lower relative to the baseline. As motor heating is closely related to torque magnitude and duration, these reductions are consistent with the observed thermal behavior. However, torque reduction alone does not establish whether support capability is altered.

\begin{figure}[htbp]       
  \centering                
  \includegraphics[width=0.95\linewidth]{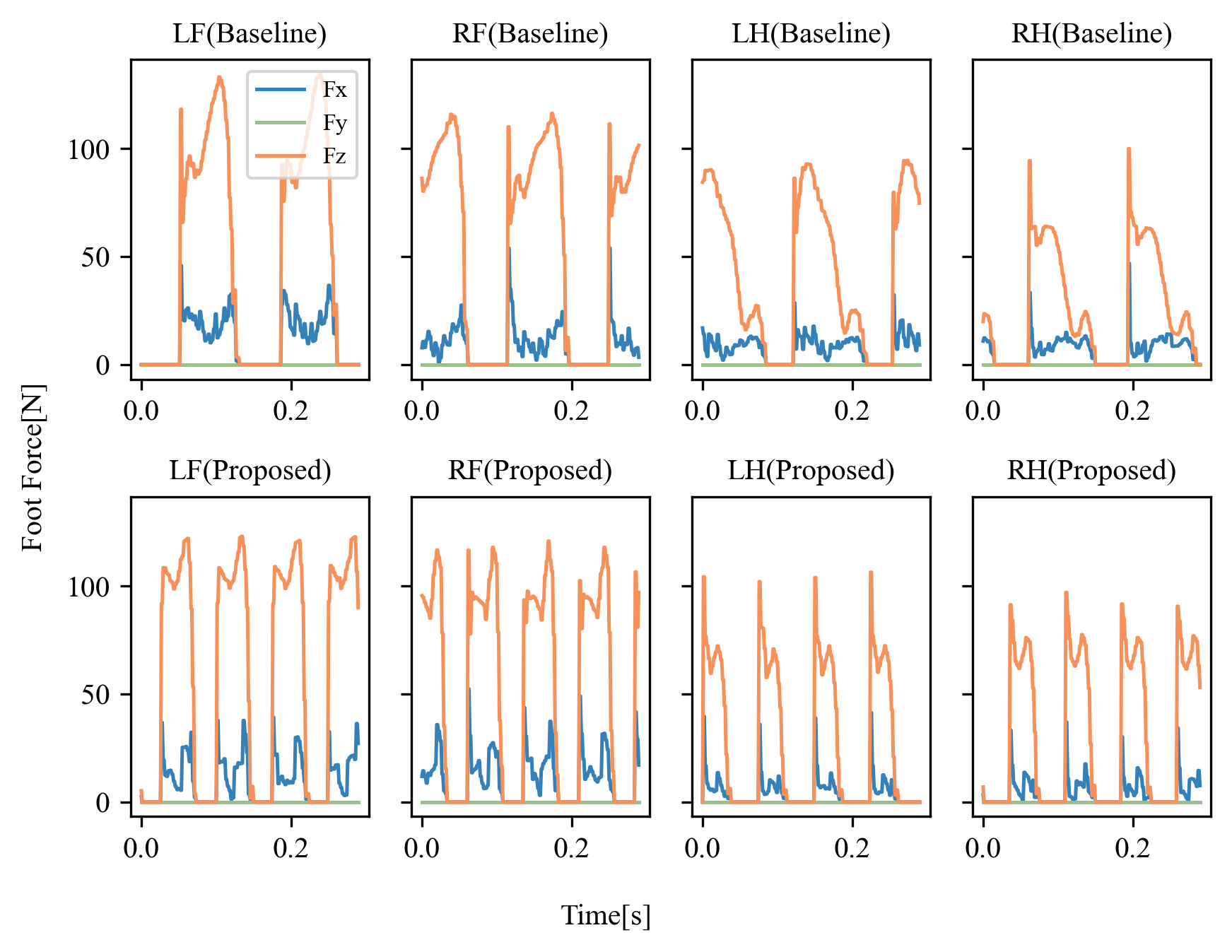}  
  \caption{Comparison of GRF curves during locomotion under the baseline and proposed policy. GRFs are expressed in a heading-aligned frame $\mathcal{H}$, obtained by rotating the world frame to align its yaw angle with that of the robot body.} 
  \label{fig:figure7}       
\end{figure}

To assess support characteristics, ground reaction forces were examined shown in Fig.\ref{fig:figure7}. Although the proposed controller exhibited a higher stepping frequency, peak vertical forces and time-averaged vertical impulse remained comparable to the baseline, while horizontal components remained relatively small. This indicates that vertical support was preserved despite reduced torque expenditure, consistent with the tracking performance observed on hardware.

\begin{figure}[htbp]       
  \centering                
  \includegraphics[width=0.95\linewidth]{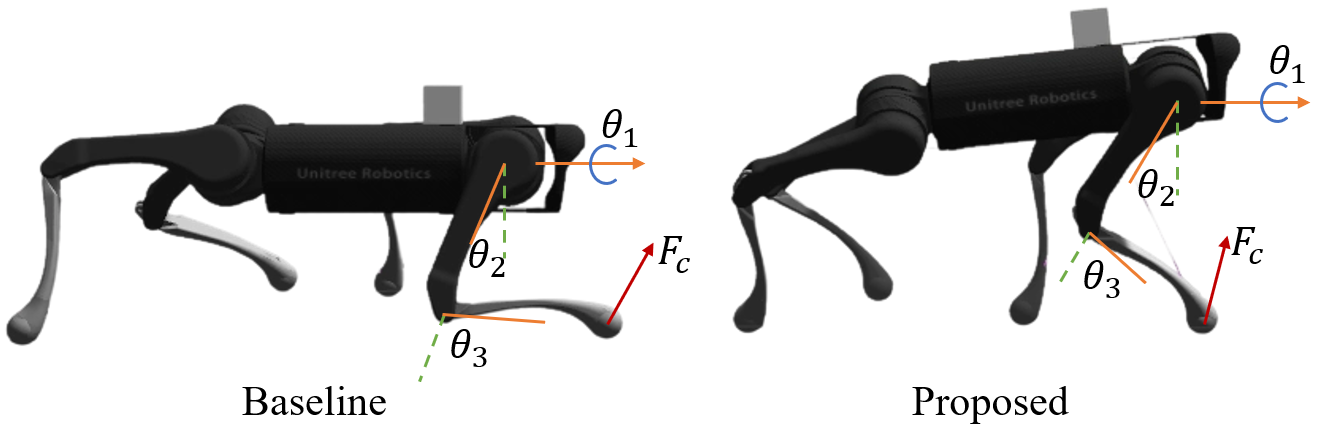}  
  \caption{Comparison of robot postures during locomotion.} 
  \label{fig:figure8}       
\end{figure}

Given that support characteristics are maintained while torque demand is reduced, the geometric mapping between joint torques and contact forces is considered. Fig.\ref{fig:figure8} shows that steady-state locomotion under the proposed controller features a greater pitch angle and a higher center of mass, implying a modified torque-to-force transmission geometry.

\begin{figure}[htbp]       
  \centering                
  \includegraphics[width=0.95\linewidth]{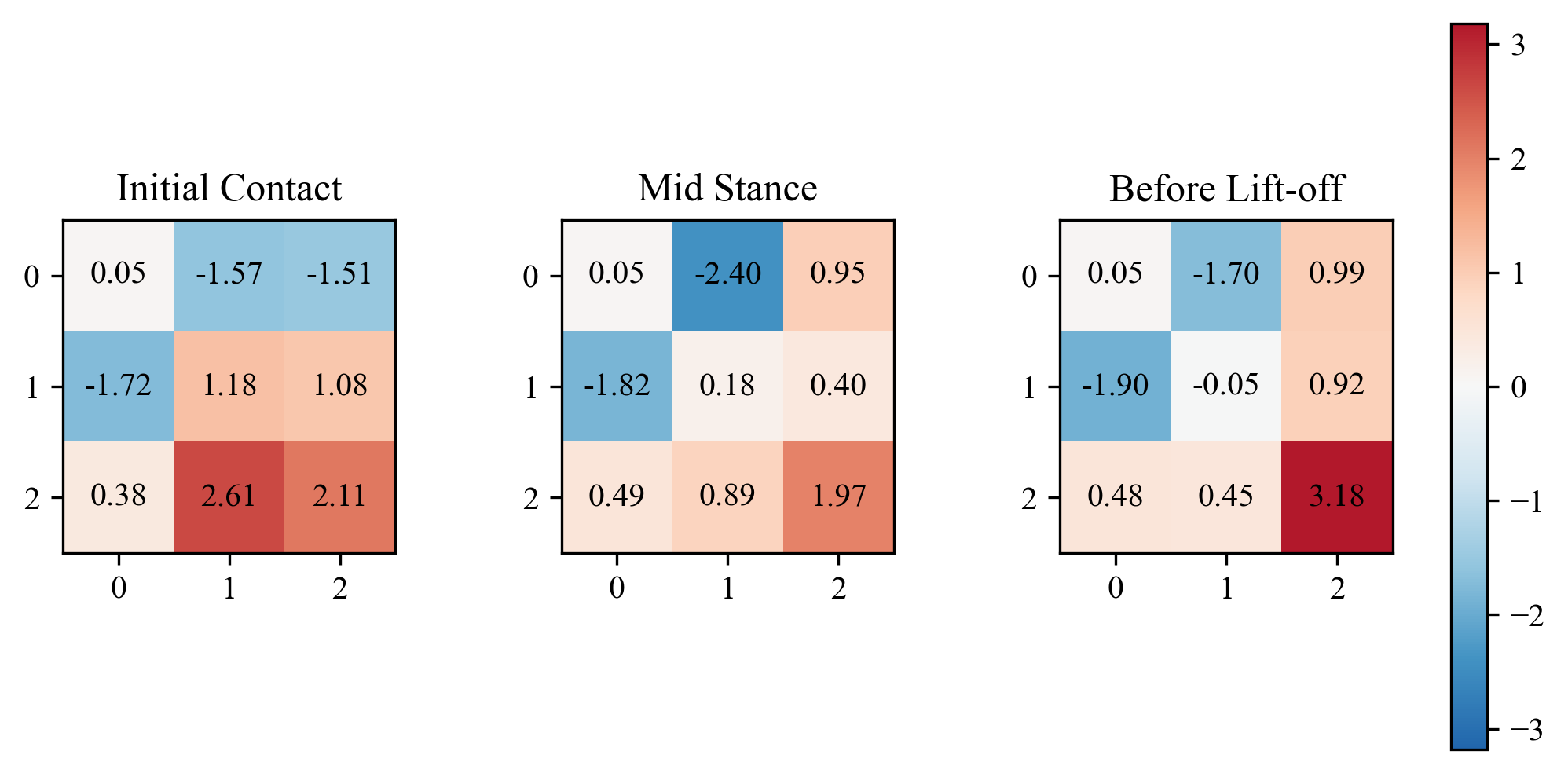}  
  \caption{Comparison of the Jacobian matrices at different time instances. The horizontal axis represents the rows of the matrix, and the vertical axis represents the columns. The values are obtained by subtracting the absolute-value Jacobian matrix under the baseline from the proposed policy.} 
  \label{fig:figure9}       
\end{figure}

This interpretation is quantitatively supported by the joint-to-foot Jacobian analysis in Fig.\ref{fig:figure9}. The absolute Jacobian difference \(|\boldsymbol{J}_{\mathrm{ours}}| - |\boldsymbol{J}_{\mathrm{base}}|\) is computed at three representative phases of stance. Positive values indicate that the same joint torque produces greater force in the associated direction. The vertical (\(z\)) components were consistently positive, whereas horizontal components were slightly reduced. Combined with the dominance of vertical support forces during trotting locomotion \((F_z \gg F_x, F_y)\), this indicates enhanced vertical force transmission efficiency.

These results collectively demonstrate that the proposed controller leveraged locomotion redundancy through posture adaptation, improved vertical transmission efficiency, and reduced torque requirements for equivalent support. This mechanism explains the reduced thermal load without degradation of tracking performance.

\section{CONCLUSION}

In this work, we propose a novel approach for motor thermal management during quadrupedal robot locomotion. We first establish a full-body thermal model of the quadruped robot and integrate it into the Isaac Gym simulation environment to enable real-time estimation of motor temperatures. Based on a relaxed control barrier function, we then design a reward function that penalizes motor overheating, and use reinforcement learning to train a thermal-aware locomotion policy. Finally, the proposed policy is deployed on a real robot and evaluated in comparison with a baseline method. Under a 3 kg payload, the robot walks continuously and stably for extended periods while keeping all motor temperatures below the threshold. Furthermore, we conduct a comparative analysis of the robot’s motion states and mechanical characteristics under ours policy and the baseline to provide insights into the underlying mechanism through which our approach achieves motor heat management.

Despite these promising results, there remains substantial room for improvement and future research. The current policy tends to adopt conservative walking postures even when motor temperatures are low, which limits the robot’s mobility and hinders performance on challenging terrains such as steep slopes or stairs. In future work, we plan to incorporate multi-mode recognition, enabling the robot to adapt its behavior based on motor temperature: maximizing locomotion performance under low-temperature conditions while prioritizing motor thermal safety under high-temperature conditions, thereby achieving a balance between performance and safety.

\section*{ACKNOWLEDGMENT}

Acknowledgements This work was partially supported by the National Natural Science Foundation of China (No. 52375014), Guangdong Innovative and Entrepreneurial Research Team Program (No. 2019ZT08Z780), and Dongguan Introduction Program of Leading Innovative and Entrepreneurial Talents (No. 20181220).


\vspace{12pt}

\end{document}